\documentclass[runningheads]{llncs}
\usepackage[T1]{fontenc}
\usepackage{graphicx}
\usepackage{booktabs}
\usepackage{cite}
\usepackage{multirow}
\usepackage{enumitem}

\usepackage{amsfonts}
\usepackage{amsmath}
\usepackage{array}

\usepackage{ulem}
\usepackage{cleveref}
\begin{document}

\title{Enhancing Robustness to Noise Corruption for Point Cloud Recognition via Spatial Sorting and Set-Mixing Aggregation Module}
\titlerunning{Spatial Sorting Mixer for Robust Point Cloud Recognition}

\author{Dingxin Zhang\inst{1} \and
Jianhui Yu\inst{1} \and
Tengfei Xue\inst{1} \and
Chaoyi Zhang\inst{1} \and
Dongnan Liu\inst{1} \and
Weidong Cai\inst{1}}
\authorrunning{D.~Zhang et al.}

\institute{School of Computer Science, University of Sydney, NSW 2006, Australia\\
\email{\{dzha2344, jianhui.yu, txue4133, chaoyi.zhang, dongnan.liu, tom.cai\}@sydney.edu.au}
}

\maketitle              
\begin{abstract}
Current models for point cloud recognition demonstrate promising performance on synthetic datasets. 
However, real-world point cloud data inevitably contains noise, impacting model robustness. 
While recent efforts focus on enhancing robustness through various strategies, there still remains a gap in comprehensive analyzes from the standpoint of network architecture design.
Unlike traditional methods that rely on generic techniques, our approach optimizes model robustness to noise corruption through network architecture design.
Inspired by the token-mixing technique applied in 2D images, we propose \textit{Set-Mixer}, a noise-robust aggregation module which facilitates communication among all points to extract geometric shape information and mitigating the influence of individual noise points. 
A sorting strategy is designed to enable our module to be invariant to point permutation, which also tackles the unordered structure of point cloud and introduces consistent relative spatial information.
Experiments conducted on ModelNet40-C indicate that Set-Mixer significantly enhances the model performance on noisy point clouds, underscoring its potential to advance real-world applicability in 3D recognition and perception tasks.
\keywords{Point cloud  \and Noise robustness \and Object recognition.}
\end{abstract}
\section{Introduction}
\label{sec:intro}

As a readily available and common form of 3D data, point clouds have garnered significant attention in the field of 3D recognition and perception~\cite{Pointnet}.
Current point-based methods have achieved promising performance on synthetic point cloud datasets~\cite{Pointnet2, DGCNN, PCT, point_mixer}. 
However, unlike the clean synthetic data, real-world point cloud data inevitably includes noise, which is caused by deviations in scanning sensors~\cite{sensor_noise} during the data collection, or by data processing~\cite{compress_noise, render_noise}.
Representative methods~\cite{Pointnet, Pointnet2, DGCNN, RSCNN, PCT, Pointmlp, Simpleview} for point cloud recognition are vulnerable to noise, due to insufficient consideration of robustness during their design and evaluation. 

\begin{figure}[h]
\begin{center}
  \includegraphics[width=1.0\linewidth]{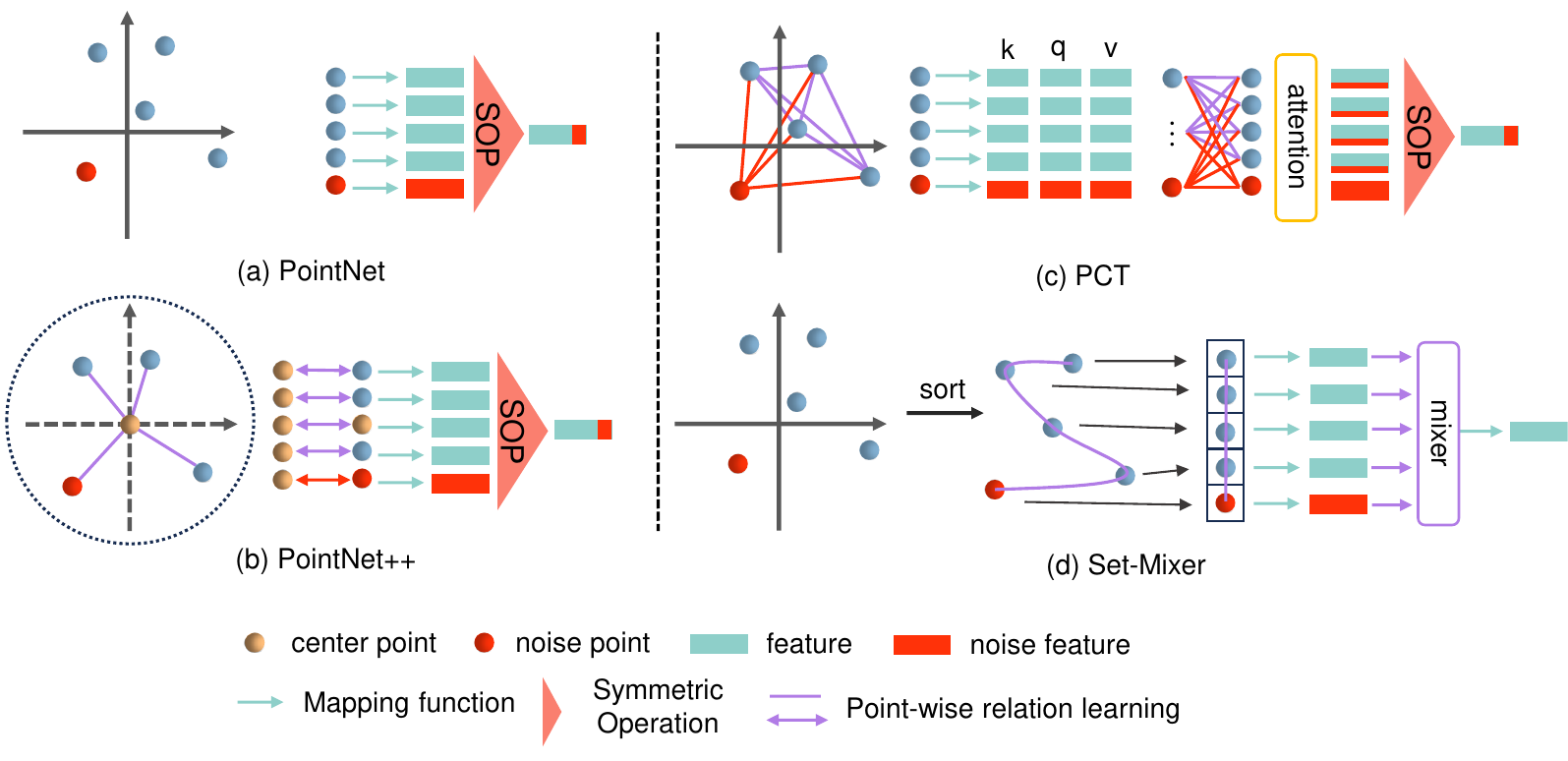}
\end{center}
  \caption{Overview of the feature extraction processes of PointNet~\cite{Pointnet}, PointNet++~\cite{Pointnet2}, PCT~\cite{PCT}, and our Set-Mixer. The red lines and squares represent the noisy relations and features. For clarity, noisy features are depicted on the far right for illustration, and they do not correspond to any specific channel.
}
\label{fig:1}
\end{figure}

Currently, researchers are increasingly focusing on improving the robustness of the model when applied in the real-world environment.
\cite{modelnet40_c} introduced a 3D corruption benchmark and generated a corrupted test dataset, ModelNet40-C, in which corruption types are classified into three categories: density, noise, and transformation.
Recent efforts aim to improve robustness through adversarial training~\cite{Dup-net, FSGM, Ada3diff}, test-time adaptation~\cite{bn, tent}, and augmentation~\cite{pointmixup, pointcutmix, rsmix}.
Adversarial training methods, primarily designed to fortify robustness against targeted attacks, have proven inadequate in handling natural noise~\cite{modelnet40_c, pointnet_c}. 
Although test-time adaptation contributes to overall robustness improvement, its drawback lies in increased inference consumption. 
Augmentation methods applied in~\cite{pointmixup, pointcutmix, rsmix, modelnet40_c} emerge as a better way to improve the robustness to noise. 

However, due to the unique nature of point clouds as sparse 3D data, the characteristics of different types of corruption vary significantly.
If the objective is to enhance robustness uniformly against all types of corruption, it becomes challenging for researchers to develop effective methods that simultaneously account for all these characteristics. Therefore, the focus often shifts to improving the generalization performance of the model.
These approaches are valuable, but by treating specific types of corruption as distinct areas of study, we can develop specialized methods based on their unique characteristics. 
For instance, the fields of rotation-invariant~\cite{RIconv, RethinkRot, Parot, sgmnet, RIconv2, LCTrans} and LiDAR recognition\cite{lidar1, lidar2, lidar3, lidar4} have already seen substantial work and have become well-developed subfields. 
Methods developed by examining the nature of point cloud corruption can deepen our understanding of point cloud data and theoretically advance the entire field of point cloud recognition.

We believe that existing methods tend to fall short when confronted with substantial noise corruption challenges. We aim to address these challenges by improving robustness through the optimization of the model structure itself, starting from the nature of noise corruption. This approach can deepen our understanding of point cloud noise corruption and, in conjunction with augmentation and training techniques, further achieve better robustness.

Existing state-of-the-art (SOTA) models~\cite{Pointnet, Pointnet2, RSCNN, PCT} for point cloud recognition exhibit noticeable deficiencies when confronted with noisy points. 
In our perspective, the prevalent aggregation modes in point cloud models tend to overemphasize the spatial positional information of single points which amplify the impact of noise points.
Take PointNet shown in Fig.~\ref{fig:1}~{(a)}, as an instance, it maps the coordinate information of each point to a high-dimensional space and then uses max pooling to aggregate the positional information of all points into a single shape feature for the entire point cloud. 
It is evident that noise can significantly affect the spatial positional information of individual points, causing abnormal features to overshadow the normal features of all other points. 
Subsequent methods, whether through gathering local groups or calculating attention weights, have not been able to avoid this issue, as shown in Fig.~\ref{fig:1}.
However, from an overall perspective, although the spatial positional information of individual points may vary significantly, the point set can still maintain a general geometric shape. 
We aim to develop an aggregation module that focus on more stable patch-wise geometric shape information, thereby enhancing the robustness of point clouds against noise.

We propose using a mixer architecture for implementing noise-robust point set feature aggregation. PointMixer~\cite{point_mixer} has applied the mixer architecture to point clouds by incorporating a positional encoding module and replacing the token-mixing module with normalization operations to improve efficiency. 
In the original MLP-Mixer~\cite{MLP_mixer}, the token-mixing module relies on token order for communication between tokens. 
However, due to the unordered nature of point clouds, this approach is ineffective in PointMixer. 
Inspired by this, we hypothesize that if we can effectively sort the points, embedding spatial information implicitly into the order, we can restore the functionality of the token-mixing module. 
Then, the mixer structure will comprehensively consider the local context of the entire point set and the relationships among all points to encode robust geometrical shape information. 
This would mitigate the impact of noisy points, extract more robust shape features of the point set, thereby enhancing the model's robustness.
We analyzed three spatial-aware sorting algorithms: Axis Projection Sorting (APS), Plane Clockwise Sorting (PCS), and Euclidean Distance Sorting (EDS). 
Additionally, we used the spatial point set center to represent the point set for sorting and gathering at deeper levels of the hierarchical network, aiming to enhance the implicit spatial information and the stability of the sorted order.

In this paper, we propose a noise-robust aggregation module, Set-Mixer, to address the challenges posed by noisy point clouds. 
Specifically, we eschew the use of symmetric functions and instead employ spatial sorting methods to handle unordered data structures, introducing consistent relative positional information among all points. 
A mixer layer is utilized for encoding the shape information of the point set.

The contributions of this work are summarized as follows: 
\begin{itemize}[itemsep=0pt, parsep=1pt]
    \item We design the Set-Mixer module for point cloud feature aggregation, which could significantly improve the robustness of models to noise corruptions.
    \item We introduce a spatial sorting strategy for Set-Mixer, which implicitly embeds relative positional information and addresses the unordered nature of point clouds, and test three sorting algorithms.
    \item Through comprehensive experiments, we prove the efficacy of our design over existing techniques on the noisy point cloud dataset (ModelNet40-C).
\end{itemize}

\section{Related works}
\label{sec:rel_work}
\subsection{Deep Learning on Point Clouds}
Recent methods for point cloud understanding have primarily concentrated on point-based approaches, driven by their advantages of low memory consumption, no information loss during preparation, and high versatility across various tasks.
The pioneering work PointNet~\cite{Pointnet} first introduced the utilization of symmetric functions for point cloud feature aggregation, thereby achieving permutation invariance to effectively handle the unordered nature inherent in point clouds.
PointNet++~\cite{Pointnet2} utilizes a grouping and aggregation structure to build a hierarchical extractor, which constructs local reference frames to extract local information. Subsequent studies~\cite{RSCNN, DGCNN} have predominantly adopted the strategy of constructing a local point set to enhance the network's learning of local features. Some recent methods, such as PCT~\cite{PCT}, Point Transformer~\cite{point_transformer}, Point Transformer V2~\cite{point_transformer_v2}, CureveNet~\cite{Curvenet} and 3DMedPT~\cite{3DMedPT} employ the transformer structure to build point-wise relations between all points during feature aggregation and enhance global feature learning.

The MLP-Mixer~\cite{MLP_mixer} is a 2D image classification model, constructed exclusively with multi-layer perceptrons (MLPs). 
It utilizes a channel-mixing module applied to image patches for local pattern learning and a token-mixing module applied across patches to capture relative spatial information.
Inspired by MLP-Mixer, PointMixer~\cite{point_mixer} devises a universal operator for point sets with a mixer structure.
The network has excluded the usage of the token-mixing, since the unordered nature of the point cloud prevents it from providing necessary spatial information for token-mixing.
As an alternative, relative position information between each point and the center point is embedded to enhance performance. 
In contrast to PointMixer, the sorting strategy implemented in our Set-Mixer closely adheres to the foundational principles of token-mixing.

\begin{figure*}[t]
\begin{center}
  \includegraphics[width=1.0\linewidth]{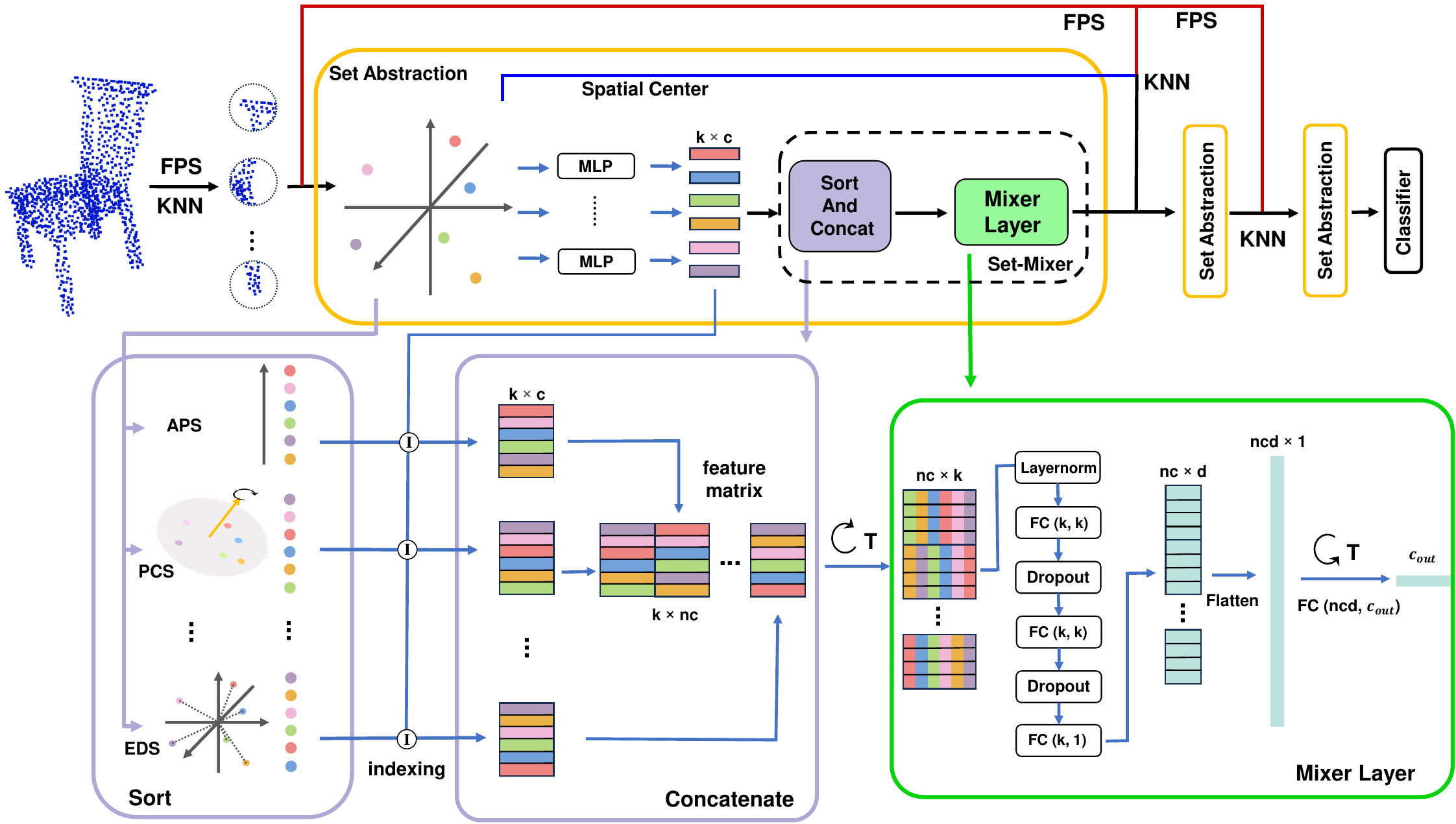}
\end{center}
  \caption{
  Illustration of the proposed Set-Mixer architecture. 
  We incorporated the Set-Mixer module into the PointNet++ structure, replacing the max-pooling function. 
  For local point sets generated with FPS and KNN, we conducted sorting operations based on coordinate values for n times, indexing and concatenating corresponding features. 
  The resulting feature matrix was then transposed and passed through the MLP Mixer layer, facilitating channel-wise cross-point learning and feature aggregation. 
  Spatial centers of point sets are calculated for KNN grouping and sorting in the next level.
  }
\label{fig:main}
\end{figure*}

\subsection{Point Cloud Model Robustness}
Extensive efforts have been invested in improving the robustness of point cloud classifiers by data augmentation, training strategy or test-time adaptation strategy.
Xiang et al.~\cite{adv_attack} first demonstrated the susceptibility of 3D machine learning models to adversarial attacks, prompting the exploration of numerous approaches aimed at enhancing attack performance and identifying potential defense mechanisms~\cite{Dup-net, FSGM, Ada3diff, adv_1, adv_2, adv_3}. 
However, as emphasized by Sun et al.~\cite{modelnet40_c}, these studies primarily focus on fortifying robustness against targeted attacks, rather than the natural noise.

Sun et al.\cite{modelnet40_c} introduced a comprehensive benchmark, ModelNet40-C, to evaluate the robustness of common 3D point cloud corruptions, including noise corruptions. 
This benchmark assessed the robustness of various representative point cloud models and the effectiveness of methods aimed at enhancing robustness. 
Data augmentation techniques, such as PointCutMix-R, PointCutMix-K\cite{pointcutmix}, PointMixup~\cite{pointmixup}, and RSMix~\cite{rsmix}, achieved the best performance. 
Additionally, test-time adaptation methods~\cite{bn, tent} demonstrated significant improvements for specific categories of corruption. 
Our proposed Set-Mixer module can be integrated with these data augmentation and test-time adaptation methods to further enhance robustness against noise.
Ren et al.~\cite{modelnet_c} presented another benchmark PointCloud-C and proposed a Robust Point cloud Classifier (RPC).
Building on these two benchmarks, recent works have achieved promising performance. For instance, CSI\cite{csi} employs density-aware sampling and an entropy minimization objective to enhance robustness. EPiC~\cite{EPiC} develops an ensemble framework based on partial point cloud sampling. Refocusing~\cite{focusing} enhances robustness by cropping the most influential points to filter out the impact of outliers.

\section{Method}
\label{sec:method}
\subsection{Point-based Network Analysis}
\label{sec:pbn_analysis}
The point cloud feature extractor, illustrated in Fig.~\ref{fig:1}, comprises of three components: a shared point-wise mapping function, geometric relation learning, and a feature aggregation module.
The pioneering PointNet structure depicted in Fig.~\ref{fig:1}(a) does not contain the relation learning component. 
Taking a point set $\mathbf{P} \in \mathbb{R}^{N \times (3 + c)}$ with $N$ points as an example, with each point $x$ constructed with its xyz position $\mathbf{p} \in \mathbb{R}^{1 \times 3}$ and corresponding $c$ channel feature $\mathbf{f} \in \mathbb{R}^{1 \times c}$,
PointNet can be formulated as:
\begin{equation} \label{eq:}
\mathbf{f}_{P} = \mathcal{A}(\{\mathcal{T}(\mathbf{p}_{x_i}, \mathbf{f}_{x_i}), \forall x_i \in \mathbf{P}\}),
\end{equation}
where $\mathcal{T}$ is the mapping function and $\mathcal{A}$ determines the max-pooling function for feature aggregation. 
Sophisticated methods explore relation learning, 
take center point $cp$ into consideration and calculate hand-crafted relative features with the function $\mathcal{R}$ to enrich geometric information: 
\begin{equation} \label{eq:center_point}
\mathbf{f}_{R} = \mathcal{R}((\mathbf{p}_{x_i}, \mathbf{f}_{x_i}), (\mathbf{p}_{c}, \mathbf{f}_{c})), \forall x_i \in \mathbf{P} .
\end{equation}
Thus, the feature extraction of PointNet++ (Fig.~\ref{fig:1}(b)) can be formulated as:
\begin{equation} \label{eq:center_point}
\mathbf{f}_{P} = \mathcal{A}(\mathcal{T}(\mathbf{f}_{R})).
\end{equation}
Here, PointNet++ subtracts point coordinates and concatenates with the point feature. Although other methods may use different strategies, the presence of noise information will not be affected.

PCT (Fig.~\ref{fig:1}(c)) further utilizes a transformer structure that incorporates relative relation between any two points within the set through the attention mechanism $\mathcal{A}tt$:
\begin{equation} \label{eq:tranformer_point}
\mathbf{f}_{P} = \mathcal{A}(\mathcal{T}(\mathbf{f}_{R}) \times \mathcal{A}tt(\mathbf{f}_{R}) ).
\end{equation}
Here, we have omitted the shortcut, offset operations and multiple attention layers for clarity.
The relation of all pairs of points is learned independently when calculating the attention weights and aggregated using the summation operation within the matrix multiplication.
Besides, the transformer structure maintains the input and output sizes, without performing aggregation. 
Therefore, PCT also employs max-pooling and mean-pooling and concatenates the outputs for aggregation at the end of the structure.

In traditional models, $\mathcal{A}$ is conventionally constrained to be a symmetry operation to solve the disorder problem of point cloud and most methods selected max-pooling as the PointNet suggested.
Max-pooling can be considered as directly concatenating the independent spatial information of all points by selecting representative features, thereby serving as the shape information for the point set.
However, this causes max-pooling to have a tendency to discard a substantial portion of point feature values, retaining only the extreme values associated with individual points.
As a result, the introduction of noisy points is likely to produce extreme values after mapping and introduce significant perturbations to the pooled feature values.
The common relative feature function $\mathcal{R}$, establishing the connection between two points through hand-crafted processes, will not affect this process.
Similarly, the structure of PCT encounters analogous issue.
Although the attention mechanism effectively extracts pair-wise relations and summarizes them for each point through matrix multiplication, the computation of Key, Value, and Query, as well as the weighting process, is susceptible to noise.
When noise is present in one of the points, it can exert a dominant influence over the acquired relation feature. 
Additionally, the transformer's structure in PCT is only used for capturing high-level features, while traditional structures are employed for capturing local features.

In this paper, we propose an MLP-based order-sensitive module, amalgamating the functions of both spatial relation learning and shape feature aggregation, named Set-Mixer. 
As illustrated in Fig.~\ref{fig:1}(d), the challenge of maintaining permutation invariance is addressed through spatial-aware point sorting, and an MLP is employed for feature aggregation while simultaneously constructing relative spatial information among all points.
Through spatial-aware sorting, we integrate spatial information into the order of the entire point set, rather than embedding point-wise position information. 
Therefore, the proposed module directly learns relations among all points and employs non-symmetric aggregation, which effectively diminishes the impact of a minor number of noisy points.

\subsection{Set-Mixer Module Structure}
\label{sec:mixer_structure}
Our Set-Mixer module design draws inspiration from the structure of the token-mixing MLPs in the MLP-Mixer model used for 2D image processing, and we address issues arising from differences in data types.
In MLP-Mixer, the image is segmented into patches with fixed positions of $S \times S$. 
The token-mixing module takes the feature map $\mathbf{f_{in}} \in \mathbb{R}^{(S \times S) \times c_{in}}$ as input and employs an MLP to facilitate channel-wise communication across all patches:
\begin{equation} \label{eq:}
\mathbf{f_{out}} = (\mathcal{M}(\mathbf{f_{in}} ^\top))^\top.
\end{equation}
The input feature must be initially transposed, in order to ensure the mixing layer has been applied to the correct token dimension and transposed back at the end.

The effectiveness of token-mixing relies on the inherent characteristics of 2D images and is intricately linked with the strategy employed for patch generation.
The images are systematically partitioned into a grid structure of fixed dimensions, typically denoted as $S \times S$, with a predetermined sequence.
This configuration renders image patches sensitive to order and maintains constant spatial locations, thereby ensuring the data structure inherently incorporates abundant spatial information.

In contrast, points in a point cloud exhibit an unordered nature and can occupy arbitrary positions within the expansive 3D space, resulting in a lack of inherent spatial relations.
Point-Mixer~\cite{point_mixer} first implements the concept of MLP-Mixer for point cloud understanding.
However, the solution for this issue is abandoning the token-mixing structure and employing the relative positional encoding scheme of PointNet++.
We choose a more robust way for this issue, by artificially constructing spatial relations with sorting strategies.
The entirety of the point cloud undergoes a sorting process, where spatial information is intricately woven into a predetermined order. 
This order is then forcibly integrated into a one-dimensional grid structure, ensuring a cohesive representation of spatial relationships within the dataset.
However, the spatial information generated for point clouds in this manner is still inferior to the spatial information of images.
To enrich spatial information, the mixer can conduct more than the multiple sorting methods respectively, and concatenate corresponding features based on the sorted index:
\begin{equation} \label{eq:our_method}
\mathbf{f}_{P} = \mathcal{M}(\mathcal{O}(\{\mathcal{T}(\mathbf{p}_{x_i}, \mathbf{f}_{x_i})\})), \forall x_i \in \mathbf{P},
\end{equation}
\begin{equation} \label{eq:sort}
\mathbf{f}_{\mathcal{O}}= \mathcal{O}(\mathbf{p}, \mathbf{f}_{\mathcal{T}})= \left[\mathcal{I}(\mathcal{S}(\mathbf{p}, t), \mathbf{f}_{\mathcal{T}})\right],
\end{equation}
where $\mathcal{M}$ denotes the mixer network and $\mathcal{O}$ signifies the ordered feature generating process. 
$\mathcal{S}$ sorts point sets along selected sorting strategy $t$ and $\mathcal{I}$ is tasked with indexing input features in accordance with the sequence.

\subsection{Spatial Aware Sorting}
\label{sec:sorting}
The selection of an appropriate sorting strategy for building spatial relations within a point cloud significantly influences the characteristics and effectiveness of the Set-Mixer. We attempted three sorting algorithms:
\begin{figure}[t]
\begin{center}
  \includegraphics[width=0.7\linewidth]{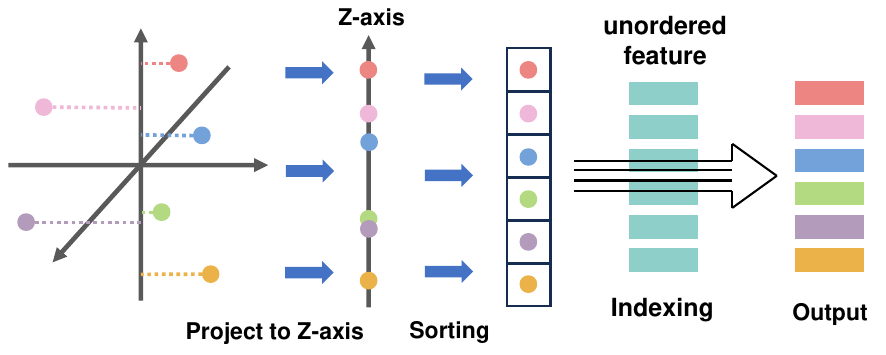}
\end{center}
  \caption{Sorting the point set along the z-axis and indexing the features, as an example of Axis Projection Sorting (APS).}
\label{fig:mixer_structure}
\end{figure}

\noindent\textbf{Axis Projection Sorting (APS)}
A straightforward way of sorting a set of 3D points $\mathbf{P}$ is directly projecting all points to an 1D axis, represented by the unit vector $e$, and then sort in the value on the axis in an ascending order:
\begin{equation} \label{eq:}
\mathbf{{APS}_{e}}(P) = {Sort}(\mathbf{p}_{x_i} \cdot e), \forall x_i \in \mathbf{P}.
\end{equation}
Fig.~\ref{fig:mixer_structure} illustrates an example of using the Z-axis as the unit vector $e$ in the APS process.

\begin{figure}[t]
\begin{center}
  \includegraphics[width=0.7\linewidth]{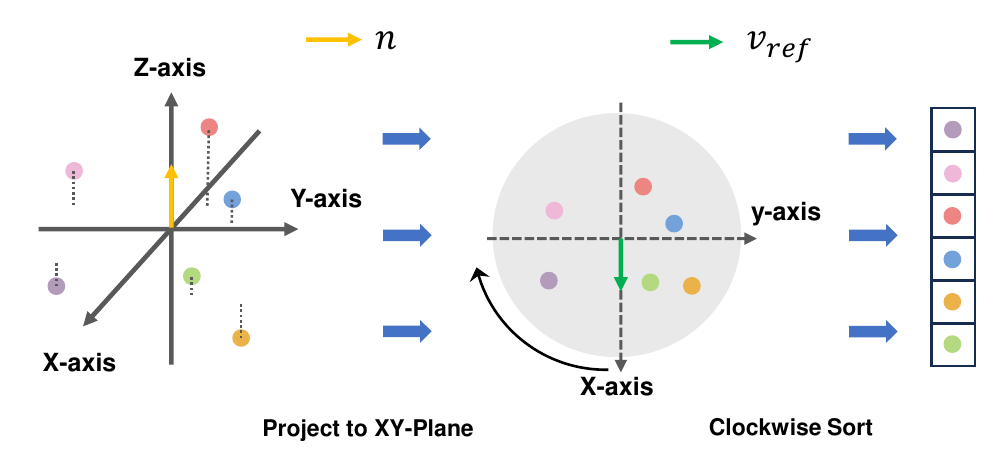}
\end{center}
  \caption{Illustration of Plane Clockwise Sorting (PCS). Z-axis and X-axis are selected as $n$ and $v_{ref}$ respectively.}
\label{fig:sort_pcs}
\end{figure}

\noindent 
\textbf{Plane Clockwise Sorting (PCS)} Zhang et al.~\cite{RIconv2} introduce a sorting operation, which is designed for generating handcrafted rotation-invariant features.
As illustrated in Fig.~\ref{fig:sort_pcs}, let $n$ denote the normal vector of a 2D plane, and $v_{ref}$ be a pre-defined reference unit vector on the plane.
The points are first projected to plane:
\begin{equation} \label{eq:}
\mathbf{P}_{proj\text{-}n} = (\mathbf{p}_{x_i} - (\mathbf{p}_{x_i} \cdot n) \cdot n), \forall x_i \in \mathbf{P},
\end{equation}
then we calculate their Cosine value corresponding to $v_{ref}$ and utilize the cross product to determine the sign of the angles:
\begin{equation} \label{eq:}
\mathbf{Angle}_{(x, v_{ref})} = (x \cdot \mathbf{v}_{ref}) \slash \|x\|),
\end{equation}
\begin{equation} \label{eq:}
\mathbf{Sign}_{(x, v_{ref})} = Sign(x \times \mathbf{v}_{ref} \cdot).
\end{equation}
Finally, we use Cosine value in range $ (-\pi, \pi] $ to sort the points in the clockwise order:
\begin{equation} \label{eq:}
\mathbf{{PCS}_{(n, v_{ref})}}(P) = {Sort}(\mathbf{Angle}_{(x_i, v_{ref})} \times  \mathbf{Sign}_{(x_i, v_{ref})}), \forall x_i \in \mathbf{P}.
\end{equation}
\noindent
\textbf{Euclidean Distance Sorting (EDS)} The third strategy is sorting points via the Euclidean distance to the center point $cp$:
\begin{equation} \label{eq:}
\mathbf{{EDS}_P} = {Sort}(\|\mathbf{p}_{x_i}-cp\|), \forall x_i \in \mathbf{P}.
\end{equation}

\subsection{Set-Mixer Network}
\label{sec:mixer_network}
As shown in Fig.~\ref{fig:main}, we built our model by replacing the max-pooling functions of PointNet++ with the proposed Set-Mixer module.
We hierarchically grouped local point sets with farthest point sampling (FPS) and k-nearest neighbours (KNN) algorithms as input.

It is noteworthy that we adjust the operation of grouping local point sets.
The original version of PointNet++ incorporates a process of generating local reference frames by subtracting the position of the query point $cp$ from each point, expressed as: $\mathbf{P}_{local} = \mathbf{P} - cp.$
It then hierarchically embeds the relative coordinates of higher levels or global coordinates of query points into the network.
This operation introduces the possibility of recurrently feeding noise information into the network, directly influencing high-level features and thus increasing the difficulty of the network in mitigating the impact of outliers.

In our proposed Set-Mixer model, we capitalize on the inherent spatial order of points, obviating the requirement for constructing local reference frames or integrating query point coordinates. Consequently, our model seamlessly employs global coordinates, circumventing the hierarchical embedding of point positions.

\noindent
\textbf{Spatial Point Set Center} 
Despite not embedding the coordinates into the network, it remains imperative to identify a central point within the point set to facilitate sorting in subsequent layers. 
Traditional point-wise methods resort to using query points as group centers due to their utilization of a subtract-and-embed-back pattern. However, in such cases, query points alone cannot effectively convey spatial positions. 
For instance, in extreme scenarios, two distinct query points positioned at the edges of a point cloud may inadvertently sample identical groups, underscoring the necessity for consistent spatial coordinates during sorting.

Consequently, we opt to compute the geometric center by averaging the coordinates of all points to serve as the spatial center:
\begin{equation} \label{eq:}
\mathbf{C}_{spatial} = \frac{1}{N}\sum_{i=1}^{N} \mathbf{P}_i.
\end{equation}
Nevertheless, the initial distribution of point cloud coordinates remains valuable. We utilize these initial coordinates for FPS sampling and querying neighbour groups, aiming to achieve more evenly distributed groups.

Assuming the input set contains $k$ points and the feature map $\mathbf{f}_{\mathcal{T}}$ has $c$ channels,
our module can be formulated as:
\begin{equation} \label{eq:mixer_structure}
\mathcal{M}(\mathbf{f}_{\mathcal{O}}) = r(m(\mathbf{f}_{\mathcal{O}}^\top)^\top).
\end{equation}
The MLP layer $m$ consists of three Fully Connected (FC) layers, facilitating channel-wise communication among all points. 
It aggregates the input feature matrix $\mathbf{f}_{\mathcal{O}} \in \mathbb{R}^{k \times 3c}$ into $\mathbf{f}_{m} \in \mathbb{R}^{d \times 3c}$, where $d$ represents the channel dimension of $m$. 
Given that the sorting process was implemented multiple times, effectively increasing the feature matrix size, an additional FC layer $r$ was employed to reduce the feature channels from $d*3*c$ to $2c$ to maintain the output feature size.
Additionally, $m$ incorporates two dropout layers, which force the network to avoid over-reliance on specific points, such as those with extreme coordinate values.
The mixing module comprehensively considers the relations among all points for aggregation, maximizing the mitigation of the impact of noisy points.

\section{Experiments and Results}

\subsection{Model Setting}

The proposed Set-Mixer module is incorporated into PointNet++~\cite{Pointnet2} structure and the Set-Mixer is capable of conducting multiple sorting operations.
To maximize the acquisition of spatial information, we utilize the three orthogonal directions x, y, and z-axes for APS sorting. Similarly, for PCS sorting, the XY-plane, YZ-plane, and XZ-plane are employed, with the x-axis, y-axis, and z-axis serving as the corresponding starting reference vector  $v_{ref}$, respectively.

\subsection{Dataset and Evaluation Criteria}
Following the ModelNet40-C~\cite{modelnet40_c} and PointCloud-C~\cite{pointnet_c} benchmark, our model was trained on the clean ModelNet40~\cite{modelnet40} dataset and employed the corrupted test sets for evaluation. Both of ModelNet40-C and PointCloud-C are derived by introducing corruption to the test set of ModelNet40. 
ModelNet40-C defines 15 distinct corruption types, with each category comprising five variations related to Density, Noise, and Transformation. Additionally, the dataset incorporates five levels of severity for each corruption type. For our evaluation, we selected Uniform, Gaussian, Impulse, Upsampling, and Background from the Noise category.
For PointCloud-C, we focused on Jitter, Drop Global/Local, and Add Global/Local, excluding Scale and Rotate as they are unrelated to noise robustness.
Besides, we also employ the real-world dataset ScanObjectNN~\cite{Scanobjectnn} (PB\_T50\_RS variant) for evaluation and created a corrupted ScanObjectNN-C set, with the method for generating corrupted samples from ModelNet40-C~\cite{modelnet40_c}.

The error rate (ER), class-wise mean error rate (mER) and Relative mean CE (RmCE), are employed as the metrics.
Following the definition of ModelNet40-C, $\mathbf{ER}^f_{s, cr}$ denotes the error rate of classifier $f$ on corruption $cr$ with severity $s$,
$\mathbf{ER}^f_{s}$ and $\mathbf{ER}^f_{\text{noise}}$ denote the mean error rate of severity $s$ and the mean error rate of 5 noise corruption types, respectively. 
RmCE is introduced by PointCloud-C, which is designed for measuring the performance drop compared to the clean test set. PointNet++ with the smallest performance drop among typical models are selected as the baseline and the metric can be formulated as:
 \begin{equation} \label{eq:}
\mathbf{RmCE} = \frac{1}{N}\sum_{i=1}^{N}  \frac{\mathbf{ER}^f_{\text{noise}} -\mathbf{ER}^f_{\text{clean}}}{\mathbf{ER}^{PointNet++}_{\text{noise}} -\mathbf{ER}^{PointNet++}_{\text{clean}}}.
\end{equation}

\subsection{Performance Analysis}
The comparison of our proposed Set-Mixer and other representative models on ModelNet40-C are illustrated in Table~\ref{table:modelnet_40c}, demonstrating the superior robustness of the Set-Mixer to noisy data. 
The proposed method exhibits the lowest error rate across all categories, particularly excelling in the Impulse and Background types, surpassing the PointNet++ by $25.4\%$ and $8.2\%$, respectively.
This is attributed to the fact that these two types of corruption introduce noisy points with extreme values. 
As previously discussed, such values will be captured by max-pooling methods, substantially contaminating the aggregated feature. 
In contrast, their impact on the Set-Mixer structure is minimal.
Furthermore, it is worth mentioning that although the proposed model exhibits a relatively higher error rate on the clean dataset, the disparity introduced by noisy corruption is significantly smaller compared to other models. 
The performance drop on the clean dataset have also been observed in CSI~\cite{csi} and CP++~\cite{focusing}. 
In our method, this issue may stem from the difficulty of restoring positional information through the current sorting strategies, and it could potentially be further refined through the implementation of more sophisticated sorting strategies for spatial information generation.

Table~\ref{table:pointcloud_c} presents the performance of the Set-Mixer on PointCloud-C and ScanObjectNN-C, together with its model size and inference computational cost.
While the Set-Mixer module increases model size, its computational cost remains low. 
This is because we utilize global coordinates, so that the computational load of the first set abstraction can be significantly reduced with code optimizations. 

\begin{table*}[t]
\centering
\caption{Classification error rate on ModelNet40-C. 
} \label{table:modelnet_40c}
\begin{tabular}{|l|c|c|c|ccccc|}
\hline
Typical Model & \multicolumn{1}{c|}{$\text{ER}_{\text{clean}}$} & \multicolumn{1}{c|}{$\text{ER}_{\text{noise}}$} & \multicolumn{1}{c|}{RmCE} & \multicolumn{1}{c}{Uniform} & \multicolumn{1}{c}{Gaus.} & \multicolumn{1}{c}{Impulse} & \multicolumn{1}{c}{Upsamp.} & \multicolumn{1}{c|}{Bg.} \\ \hline
PointNet~\cite{Pointnet} & 9.3 & 32.7 & 1.62 & 12.4 & 14.4 & 29.1 & 14.0 & 93.6 \\
PointNet++~\cite{Pointnet2} & 7.0 & 21.5 & 1.00 & 20.4 & 16.4 & 35.1 & 17.2 & 18.6  \\
DGCNN~\cite{DGCNN} & 7.4 & 25.7 & 1.26 & 14.6 & 16.6 & 24.9 & 19.1 & 53.1 \\
RSCNN~\cite{RSCNN} & 7.7 & 25.5 & 1.23 & 24.6 & 18.3 & 46.2 & 20.1 & 18.3 \\
PCT~\cite{PCT} & 7.1 & 28.1 & 1.45 & 12.1 & 13.9 & 39.1 & 17.4 & 57.9 \\
SimpleView~\cite{Simpleview} & \textbf{6.1} & 23.6 & 1.21 & 14.5 & 14.2 & 24.6 & 17.7 & 46.8 \\
CurveNet~\cite{Curvenet}    & \uline{6.2} & 24.8 & 1.28 & 12.0	& 12.2	& 24.8 & 11.3& 63.5\\
GDANet~\cite{GDANet} & 6.6 & 20.3 & 0.94 & 12.6 & 14.6 & 22.9 & 17.3 & 34.1 \\
PAConv~\cite{PAConv}    & 6.4 & 34.1 & 1.63 & 25.9	& 29.8	& 32.0 & 27.0 & 56.7\\
PointMLP~\cite{Pointmlp} & 6.5 & 36.9 & 2.10 & 22.8 & 32.6 & 40.2 & 38.7 & 50.3 \\
PointMixer~\cite{point_mixer} & 6.8 & 28.5 & 1.50 & 16.5 & 17.9 & 29.4 & 16.6 & 62.1 
  \\\hline \hline
Robust Model & \multicolumn{1}{c|}{$\text{ER}_{\text{clean}}$} & \multicolumn{1}{c|}{$\text{ER}_{\text{noise}}$} & \multicolumn{1}{c|}{RmCE} & \multicolumn{1}{c}{Uniform} & \multicolumn{1}{c}{Gaus.} & \multicolumn{1}{c}{Impulse} & \multicolumn{1}{c}{Upsamp.} & \multicolumn{1}{c|}{Bg.} \\ \hline
EPiC-DGCNN~\cite{EPiC} & 7.0 & 11.6 & 0.32 & 10.5 & 10.9 & 11.6 &  11.6 &  13.6 \\ 
EPiC-PCT~\cite{EPiC} & 6.6 & 11.5 & 0.34 & 11.9 & 13.0 & 11.2 &  12.3 &  \textbf{9.2} \\ 
CSI-PCT~\cite{csi} & 7.3 & 11.3 & 0.28 & 10.8 & \uline{10.7} & 12.3 &  11.6 &  11.2 \\ 
CP++-DGCNN~\cite{focusing} & 8.4 & 11.1 & 0.19 & \textbf{10.5} & 10.9 & \uline{10.1} & 12.0 & 12.0\\ 
CP++-RPC~\cite{focusing} &8.8 & 13.7 & 0.34 & 11.5 & 13.3 & 14.0 & 12.8 & 17.0 \\ 
\hline 
Set-Mixer-APS & 8.1 & \textbf{10.0} & \uline{0.13} & \textbf{10.5} & \textbf{10.1} & \textbf{9.7} & \textbf{9.4} & \uline{10.4} \\ 
Set-Mixer-PCS  & 9.6 & \uline{11.0} & \textbf{0.10} & 11.1& 11.1 & 10.8 & \uline{10.8} & 11.3 \\ 
Set-Mixer-EDS  & 8.2 & 20.2 & 0.83 & 11.7& 12.2 & 24.9 & 12.7 & 39.4 \\ \hline
\end{tabular}
\end{table*}

\begin{table}[t]
\centering
\caption{Model efficiency and performance on PointCloud-C and ScanObjectNN-C.}\label{table:pointcloud_c}

\begin{tabular}{|l|c|c|c|c|c|c|}
\hline
\multirow{2}{*}{Typical Model} & \multicolumn{2}{c|}{PointCloud-C} & \multicolumn{2}{c|}{ScanObjectNN-C}& \multirow{2}{*}{Param.} & \multirow{2}{*}{FLOPs}\\ \cline{2-5} 
                               & \multicolumn{1}{c|}{$\text{ER}_{\text{clean}}$}  & {$\text{ER}_{\text{noise}}$}   &   \multicolumn{1}{c|}{$\text{ER}_{\text{clean}}$}   &  {$\text{ER}_{\text{noise}}$} & & \\ \hline
PointNet++~\cite{Pointnet2}     & 7.0 & 27.2 & 22.2 & 51.9 &  1.48M & 1.68G\\
GDANet~\cite{GDANet}         & 6.6 & 20.3 & 23.3 & 52.4  & 0.93M &  9.66G\\
PointMLP~\cite{Pointmlp}       & \textbf{6.5} & 29.7 & \textbf{15.6} & 54.0 &  13.24M &  15.67G\\
EiPC-PCT~\cite{EPiC}     & 6.6 & 16.5 & - & - &  8.82M & 10.61G\\
CP++-DGCNN~\cite{focusing}  & 8.6 & 25.2 & - & - &  1.81M & 6.88G\\ \hline\hline
Set-Mixer-APS  & 8.1 & \textbf{16.4} & 22.8 & \textbf{37.4} &  8.02M & 1.76G\\ 
Set-Mixer-PCS  & 9.6 &    16.5  & 22.8 & 37.6 & 8.02M & 1.76G \\
Set-Mixer-EDS  & 8.2 &    29.5  & 30.7 & 55.7 & 3.50M &  0.71G \\\hline
\end{tabular}

\end{table}

\begin{table}[t]
\centering
\caption{Average error rates on ModelNet40-C with different data augmentation strategies of all severity and severity 5 only. $\text{ER}_{\text{n}}$ is a shorthand notation for $\text{ER}_{\text{noise}}$.}\label{table:augmentation}
\begin{tabular}{|l|cc|cc|cc|cc|}
\hline
\multirow{2}{*}{Model} & \multicolumn{2}{c|}{Standard} & \multicolumn{2}{c|}{PointCutMix-R} & \multicolumn{2}{c|}{PointCutMix-K} & \multicolumn{2}{c|}{RSMix/WolfMix} \\ \cline{2-9}
& \multicolumn{1}{c|}{$\text{ER}_{\text{n}}$} & $\text{ER}_{5}$ & \multicolumn{1}{>{\centering\arraybackslash}p{1.05cm}|}{$\text{ER}_{\text{n}}$} & $\text{ER}_{5}$ & \multicolumn{1}{>{\centering\arraybackslash}p{1.05cm}|}{$\text{ER}_{\text{n}}$} & $\text{ER}_{5}$ & \multicolumn{1}{>{\centering\arraybackslash}p{1.2cm}|}{$\text{ER}_{\text{n}}$} & $\text{ER}_{5}$ \\ \hline
PointNet~\cite{Pointnet} & \multicolumn{1}{c|}{32.7} & 35.4 & \multicolumn{1}{c|}{18.0} & 19.4 & \multicolumn{1}{c|}{21.8} & 25.2 & \multicolumn{1}{c|}{27.3} & 29.4 \\ \hline
PointNet++~\cite{Pointnet2} & \multicolumn{1}{c|}{21.5} & 35.7 & \multicolumn{1}{c|}{12.2} & 15.0 & \multicolumn{1}{c|}{16.9} & 30.0 & \multicolumn{1}{c|}{19.3} & 41.0 \\ \hline
DGCNN~\cite{DGCNN} & \multicolumn{1}{c|}{32.6} & 36.3 & \multicolumn{1}{c|}{11.4} & 12.8 & \multicolumn{1}{c|}{11.9} & 14.8 & \multicolumn{1}{c|}{13.0} & 16.7 \\ \hline
PCT~\cite{PCT} & \multicolumn{1}{c|}{25.5} & 40.0 & \multicolumn{1}{c|}{10.5} & 11.8 & \multicolumn{1}{c|}{12.6} & 17.7 & \multicolumn{1}{c|}{12.0} & 17.0 \\ \hline
EPiC-PCT~\cite{EPiC} & \multicolumn{1}{c|}{11.5} & 17.0 & \multicolumn{1}{c|}{-} & - & \multicolumn{1}{c|}{-} & - & \multicolumn{1}{c|}{11.5} & 15.5 \\ \hline
CSI-PCT~\cite{csi}  & \multicolumn{1}{c|}{11.3} & - & \multicolumn{1}{c|}{9.6} & - & \multicolumn{1}{c|}{-} & - & \multicolumn{1}{c|}{-} & - \\ \hline
Set-Mixer-APS & \multicolumn{1}{c|}{\textbf{10.0}} & \textbf{11.4} & \multicolumn{1}{c|}{\textbf{9.0}} & \textbf{9.9} & \multicolumn{1}{c|}{\textbf{9.1}} & \textbf{11.0} & \multicolumn{1}{c|}{\textbf{9.3}} & \textbf{10.5} \\ \hline
\end{tabular}
\end{table}

\subsection{Data Augmentation Strategies}
The proposed Set-Mixer enhances robustness by optimizing the aggregation module, thereby allowing it to effectively integrate with well-established generic techniques to further improve its robustness. 
By employing augmentation strategies, at the expense of increased training time, performance can be further enhanced.
Given the remarkable efficacy of PointCutMix-R, PointCutMix-K~\cite{pointcutmix}, and RSMix~\cite{rsmix} in enhancing the robustness of conventional models, we assessed their impact when used with Set-Mixer.
As shown in Table~\ref{table:augmentation}, augmentation methods efficiently reduce the error rate of all methods, with Set-Mixer achieving the lowest error rate.
It is noteworthy that EPiC uses WolfMix~\cite{pointnet_c}, which combines PointWolf~\cite{pointwolf} with RSMix, but this additional operation does not significantly improve robustness to noise.
The error rate under severity 5 is also calculated and it shows the noise robustness advantage of Set-Mixer becomes even more pronounced as the noise level increases.

\begin{table}[t]
\centering
\caption{Classification error rate on ModelNet40-C with different dropout rates and with/without Layernorm in the Set-Mixer-APS. 
} \label{table:dropout}
\begin{tabular}{|c|c|c|c|c|ccccc|}
\hline
\multirow{2}{*}{dropout} & \multirow{2}{*}{layernorm} & \multirow{2}{*}{$\text{ER}_{\text{clean}}$} & \multirow{2}{*}{$\text{ER}_{\text{noise}}$} & \multirow{2}{*}{RmCE} & \multicolumn{5}{c|}{Noise Type}                                                                                                            \\ \cline{6-10} 
                         &                            &                        &                   &                       & \multicolumn{1}{c}{Uniform} & \multicolumn{1}{c}{Gaus.} & \multicolumn{1}{c}{Impulse} & \multicolumn{1}{c}{Upsamp.} & Bg. \\ \hline
dp 0.0                   &         \checkmark         & \textbf{8.1}                    & 10.3              &      0.15           & 10.6                         & 10.7                          & 9.8                          & 10.0                              & 10.3       \\
dp 0.1                   &         \checkmark         & 8.7                    & 10.5              &     0.12            & 10.5                         & 10.7                          & 10.0                           & 10.5                            & 10.6       \\
dp 0.2                   &         \checkmark         & \textbf{8.1}                    & \textbf{10.0}                &    0.13             & 10.5                         & 10.1                          & 9.7                          & 9.4                             & 10.4       \\
dp 0.3                   &        \checkmark          & 8.5                    & 10.2              &    0.12             & 10.1                         & 10.0                            & 9.5                          & 10.3                            & 11.1       \\ \hline
dp 0.0                   &                            & 8.8                   & 10.9               &   \textbf{0.11}              &  10.9                         & 10.9                          & 10.2                        & 10.8                       &    11.5   \\
dp 0.2                   &                            & 8.8                    & 10.5              &    0.12              & 10.7                         & 10.6                          & 9.9                          & 10.1                            & 11.2       \\ \hline
\end{tabular}
\end{table}

\subsection{Dropout and Layer Normalization}
Two dropout layers and a layer normalization layer are employed in the Set-Mixer, preventing the layer over-reliance on specific points and enhancing training stability.
To further analyze the impact of the those two operations, we conducted an ablation experiment. 
The results, presented in Table~\ref{table:dropout}, indicate that the dropout layer exerts minimal influence on performance when assessed on a clean test set but exhibits a more pronounced impact on noisy test sets.
Remarkably, the proposed model demonstrates substantial robustness to noise even in the absence of the dropout layer.
Besides, we observe that layer normalization effectively enhances accuracy on both clean and corrupted datasets. 
Nonetheless, our model still outperforms other models without layer normalization, indicating that the primary contributing factor is the Set-Mixer's architecture, which is designed to effectively extract geometric shape information.

\begin{table}[t] 
\centering

\caption{Classification error rate on ModelNet40-C with different center porting selection strategy and aggregation strategies in the Set-Mixer-APS. 
} \label{table:spatialcenter}
\begin{tabular}{|c|c|c|c|c|c|} 
\hline
      & Query Point & Spatial Center & Max Pooling & Mean Pooling & Mixer w/o sorting \\ \hline
$\text{ER}_{\text{clean}}$ & 8.5 & \textbf{8.1} & 8.5 & 12.8 & 10.6 \\ \hline
$\text{ER}_{\text{noise}}$ & 10.8 & \textbf{10.0} & 15.7 & 18.9 & 13.4\\ \hline
RmCE  & 0.16 &  \textbf{0.13} & 0.50 & 0.42 & 0.19 \\ \hline

\end{tabular}
\end{table}

\subsection{Analysis of Spatial Center and Sorting Strategy Analysis}
We conducted additional comparative experiments using Set-Mixer-APS to substantiate the efficacy of sorting based on the spatial center. As shown in Table~\ref{table:spatialcenter}, leveraging the spatial center outperforms the query point center. Remarkably, even without incorporating the spatial center, our model's sorting strategy substantially enhances robustness. 
Furthermore, we tested the model using a pooling function and the mixer without the sorting operation. The results indicate that the mixer structure is more robust, while the sorting operation effectively improves performance.

\section{Conclusion}
In this work, we introduce Set-Mixer, a robust point set aggregation module tailored to increase the robustness to noise for point cloud recognition models.
We initiate the process by sorting a mapped point-wise feature based on spatial position to rectify disorder within point clouds. 
Subsequently, an MLP layer is employed to learn spatial relations across all points and facilitate aggregation, thereby mitigating the impact of noise points on the final extracted geometric shape features.
Experimental results on ModelNet40-C demonstrate that integrating Set-Mixer into SOTA point cloud recognition models significantly improves the model's robustness to noise.

%
%
%
\bibliographystyle{splncs04}
\bibliography{mybib}

\newpage
\appendix
\section*{Appendix}
\addcontentsline{toc}{section}{Appendix} 

\renewcommand{\thefigure}{A\arabic{figure}}
\renewcommand{\thetable}{A\arabic{table}}

\section{Model Structure}
The proposed Set-Mixer module is incorporated into PointNet++ structure and the detailed structure of our model is represented using the formats $SA($num of sampled points, number of points sampled in a patch $K$, [Fully connected layer channel]$)$ and $\text{Set-Mixer}($num of input channel $c$, $K$, dropout rate$)$:
\begin{equation*} \label{eq:}
\begin{split}
    &SA(512, 32, [64, 64, 64], \text{Set-Mixer}(64, 32, 0.2)) \rightarrow \\
    &SA(128, 64, [128, 128, 128], \text{Set-Mixer}(128, 64, 0.2)) \rightarrow \\
    &SA(1, 128, [256, 512, 512], \text{Set-Mixer}(512, 128, 0.2)) \rightarrow \\
    &FC(512, 0.5)  \rightarrow FC(256, 0.5)  \rightarrow  FC(40).
\end{split}
\end{equation*}
Here, the decimal between 0 and 1 inside $FC()$ indicates the dropout rate.

\section{Sorting Strategy}
In the proposed Set-Mixer, we perform feature sorting three times independently along the xyz-axis to enrich spatial information. We also evaluated alternative sorting strategies, including sorting once or twice, with the comprehensive results illustrated in Table~\ref{table:sorting}.

As demonstrated by Table~\ref{table:sorting}, all sorting methods exhibit superior performance compared to PointNet++. Notably, conducting multiple sorting operations yields a significant improvement. In general, increasing the number of sorting operations would enrich the spatial information of point sets, leading to enhanced performance and robustness.

\begin{table*}[h]
\centering
\caption{Classification error rate on ModelNet40-C with different sorting strategies.} \label{table:sorting}
\begin{tabular}{|c|ccc|c|c|ccc|}
\hline
  & \multicolumn{3}{c|}{APS}                                                    & \multicolumn{1}{c|}{EDS}  & PCS & \multicolumn{3}{c|}{}                                       \\ \hline
Model & x-axis                   & y-axis               & \multicolumn{1}{c|}{z-axis} & \multicolumn{1}{c|}{distance} & XY-XZ-YZ plane & \multicolumn{1}{c|}{$\text{ER}_{\text{clean}}$}    & \multicolumn{1}{c|}{$\text{ER}_{\text{noise}}$}     &  RmCE  \\ \hline
A & \checkmark                      & \checkmark            & \checkmark              &                        &     & \multicolumn{1}{c|}{\textbf{8.1}}    & \multicolumn{1}{c|}{\textbf{10.0}}     &  0.13    \\ \cline{1-1} \cline{7-9} 
B & \checkmark                      & \checkmark             &                        &                        &     & \multicolumn{1}{c|}{8.8} & \multicolumn{1}{c|}{10.9} & 0.14 \\ \cline{1-1} \cline{7-9} 
C & \checkmark              &                        & \checkmark              &                        &     & \multicolumn{1}{c|}{9.5} & \multicolumn{1}{c|}{11.4} & 0.13 \\ \cline{1-1} \cline{7-9} 
D &                        & \checkmark            & \checkmark            &                        &     & \multicolumn{1}{c|}{8.7} & \multicolumn{1}{c|}{10.5} & 0.13 \\ \cline{1-1} \cline{7-9} \hline
E & \checkmark            &                        &                        &                        &     & \multicolumn{1}{c|}{9.9} & \multicolumn{1}{c|}{13.0} & 0.21 \\ \cline{1-1} \cline{7-9} 
F &                        & \checkmark             &                        &                        &     & \multicolumn{1}{c|}{9.6} & \multicolumn{1}{c|}{13.5} & 0.27 \\ \cline{1-1} \cline{7-9} 
G &                        &                        & \checkmark            &                        &     & \multicolumn{1}{c|}{9.4} & \multicolumn{1}{c|}{12.6} & 0.22 \\ \cline{1-1} \cline{7-9} \hline
H &                        &                        &                        & \checkmark           &     & \multicolumn{1}{c|}{8.2} & \multicolumn{1}{c|}{20.2} & 0.83 \\ \cline{1-1} \cline{7-9} 
I &                        &                        &                        &                        & \checkmark & \multicolumn{1}{c|}{9.6} & \multicolumn{1}{c|}{11.0} & \textbf{0.10} \\ \cline{1-1} \cline{7-9} 
J &                        &                        &                        & \checkmark             & \checkmark  & \multicolumn{1}{c|}{8.7} & \multicolumn{1}{c|}{11.3} & 0.18 \\ \cline{1-1} \cline{7-9} \hline
K & \checkmark             & \checkmark            & \checkmark            & \checkmark              &     & \multicolumn{1}{c|}{8.2} & \multicolumn{1}{c|}{10.1} & 0.13 \\ \cline{1-1} \cline{7-9} 
L & \checkmark              & \checkmark            & \checkmark              &                        & \checkmark & \multicolumn{1}{c|}{8.8} & \multicolumn{1}{c|}{10.2} & \textbf{0.10} \\  \cline{1-1}  \cline{7-9} 
M & \multicolumn{1}{c}{\checkmark} & \checkmark & \checkmark & \multicolumn{1}{c|}{\checkmark} & \checkmark & \multicolumn{1}{c|}{8.3} & \multicolumn{1}{c|}{10.3} & 0.14 \\ \hline
\end{tabular}
\end{table*}

\section{Spatial Center Visualization}
\begin{figure}[t]
\begin{center}
  \includegraphics[width=1.0\linewidth]{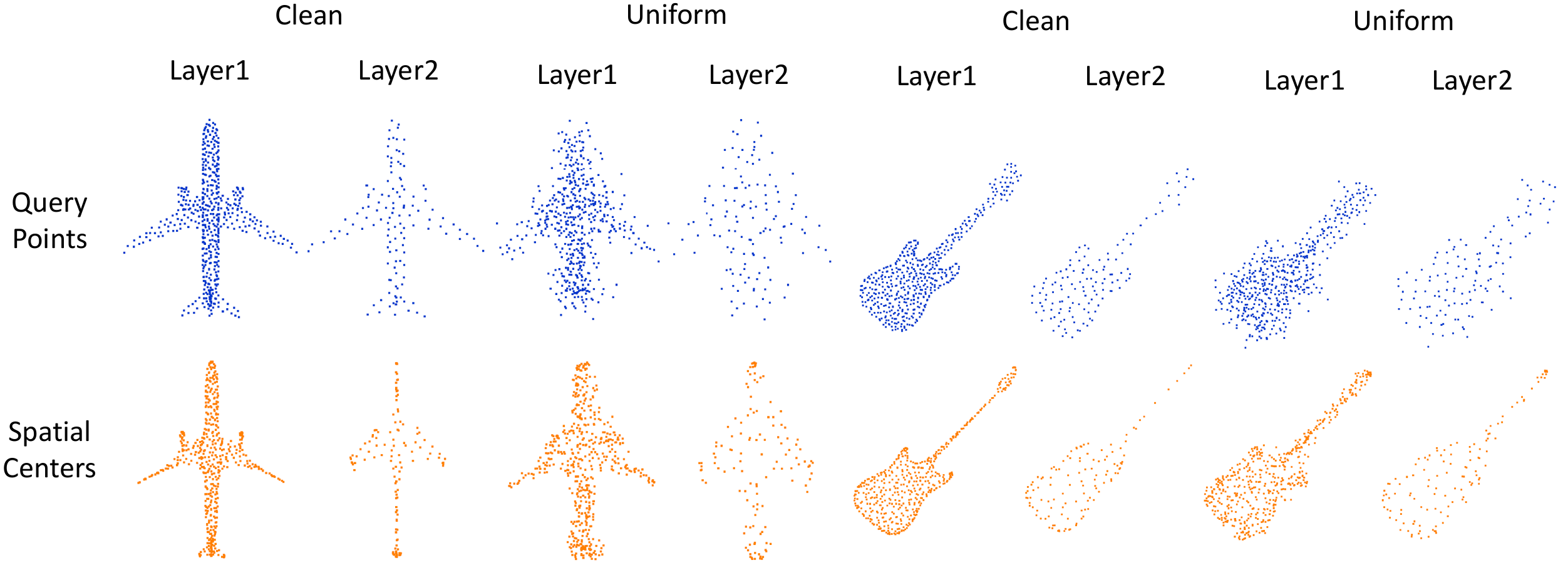}
\end{center}
  \caption{Visualization of query points and spatial center points of  layer 1 $($512 sets, 32 neighbours$)$ and layer 2 $($128 sets, 64 neighbours$)$ of Set-Mixer.  Query points are depicted in blue, while the spatial centers are marked in orange.
  }
\label{fig:center_vis}
\end{figure}

In Fig.~\ref{fig:center_vis}, we visualize both the query points and the spatial centers generated by Set-Mixer in layers 1 and 2. 
Through this visualization, we can observe that the spatial center not only provides a more effective representation of the spatial positions of point sets but also exhibits greater stability.

\section{Noisy Feature Visualization}

\begin{figure}[t]
\begin{center}
  \includegraphics[width=0.8\linewidth]{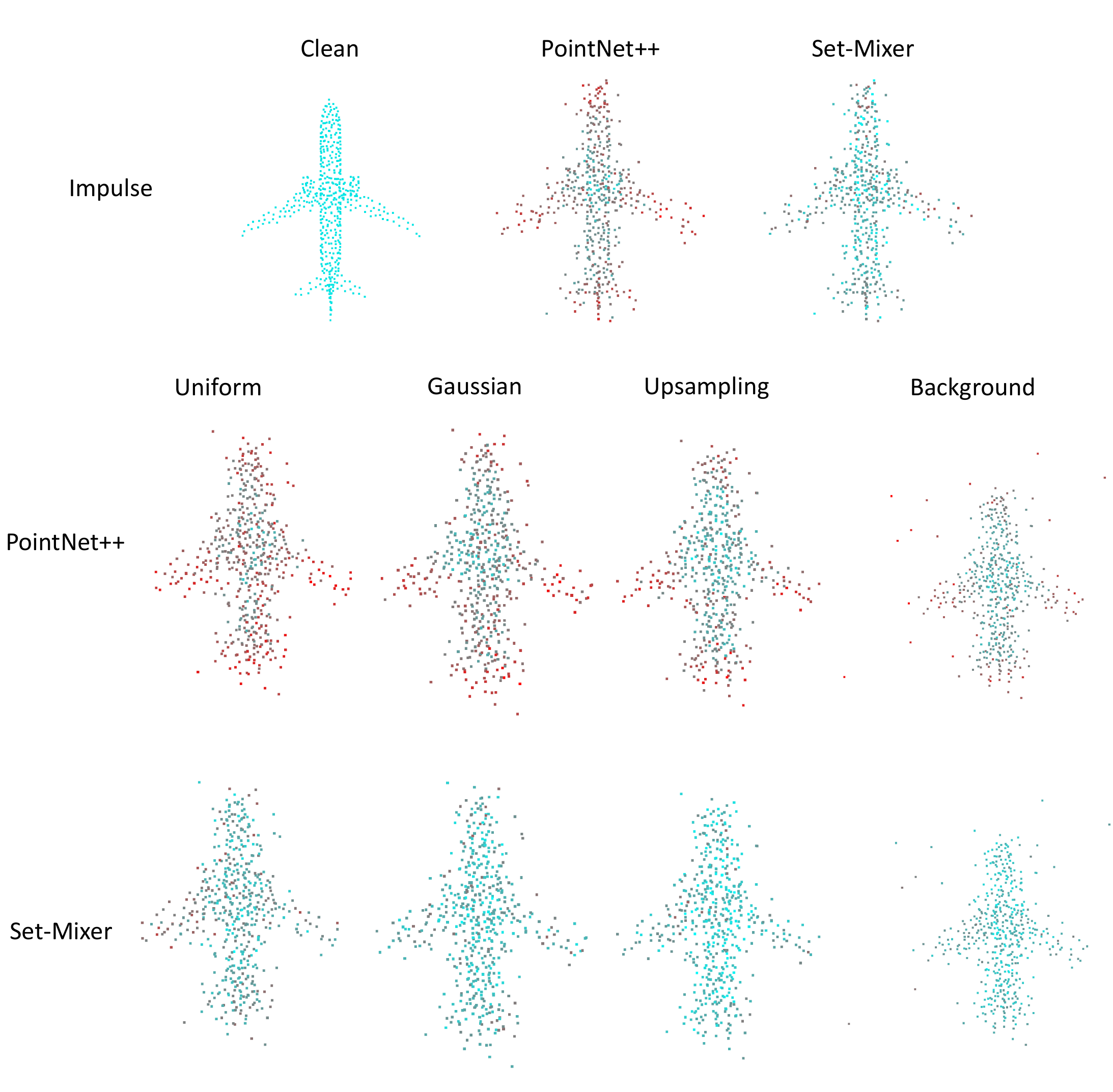}
\end{center}
  \caption{Visualization of feature change magnitudes.
  Blue indicates no change, while red indicates a high degree of change. 
  The left graph of the top line is the result of the clean sample, while the middle and right graphs illustrate the results of PointNet++ and Set-Mixer, respectively, with the impulse of severity 3.}
\label{fig:feature}
\end{figure}

\begin{figure*}[t]
\begin{center}
  \includegraphics[width=1.0\linewidth]{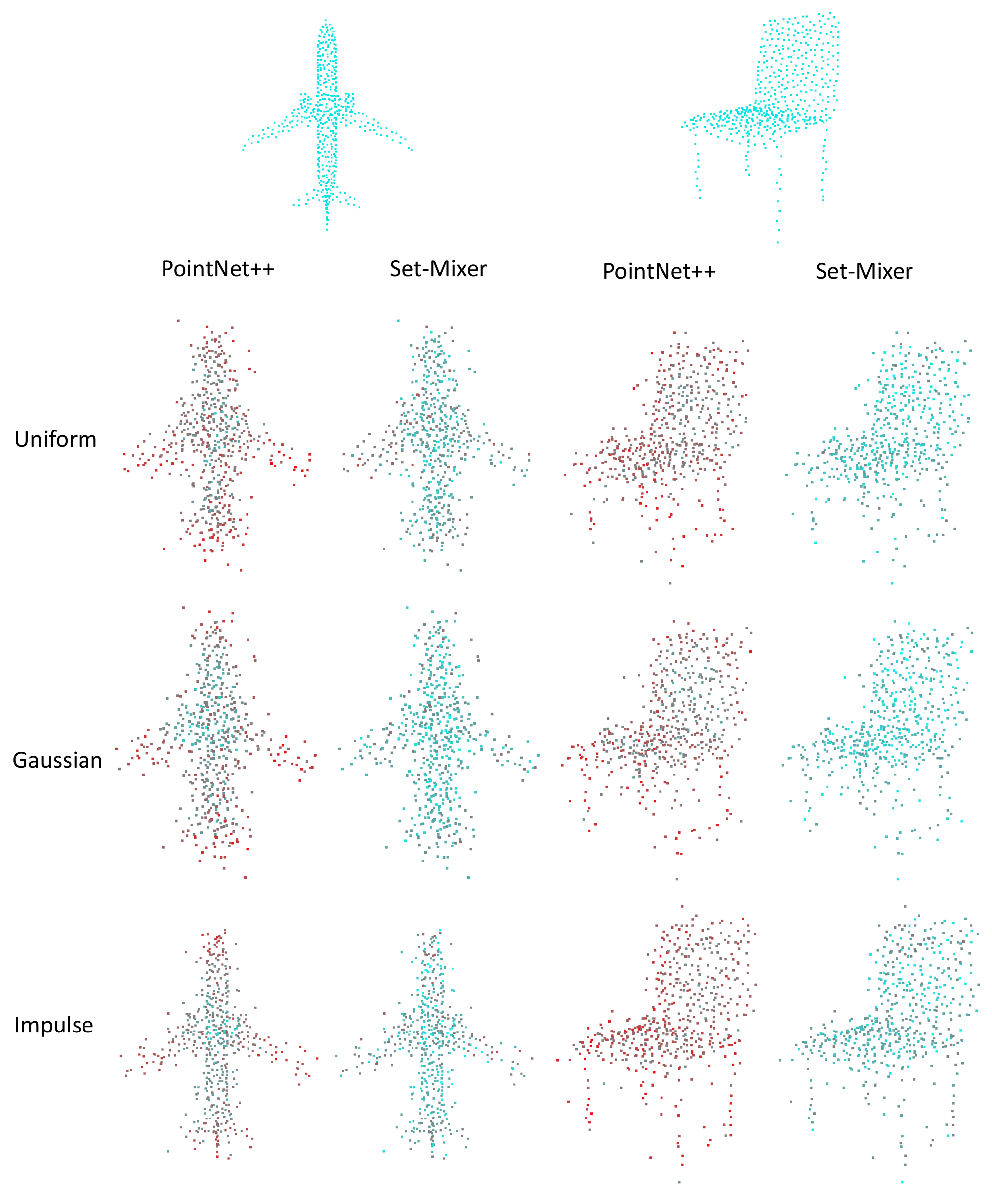}
\end{center}
  \caption{
  Visualization of feature change magnitudes compared to clean input. 
  Blue indicates no change, while red indicates a high degree of change. 
  The two graphs at the top represent the results for the original clean sample. The severity is set to 3.
  }
\label{fig:vis_page1}
\end{figure*}

\begin{figure*}[t]
\begin{center}
  \includegraphics[width=1.0\linewidth]{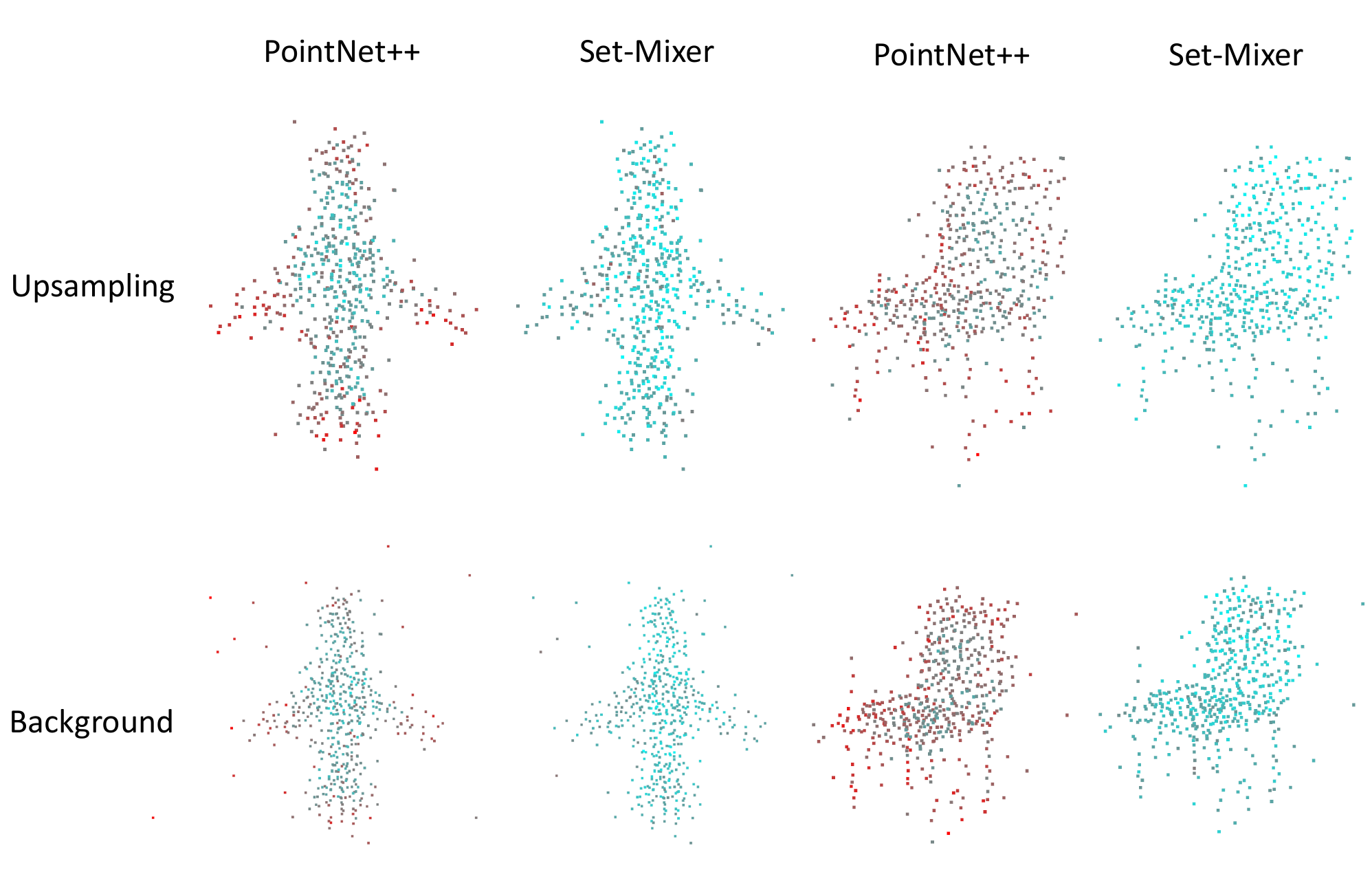}
\end{center}
  \caption{
  Visualization of feature change magnitudes compared to clean input. 
  Blue indicates no change, while red indicates a high degree of change. 
  The two graphs at the top represent the results for the original clean sample. The severity is set to 3.
  }
\label{fig:vis_page2}
\end{figure*}

\begin{figure*}[t]
\begin{center}
  \includegraphics[width=1.0\linewidth]{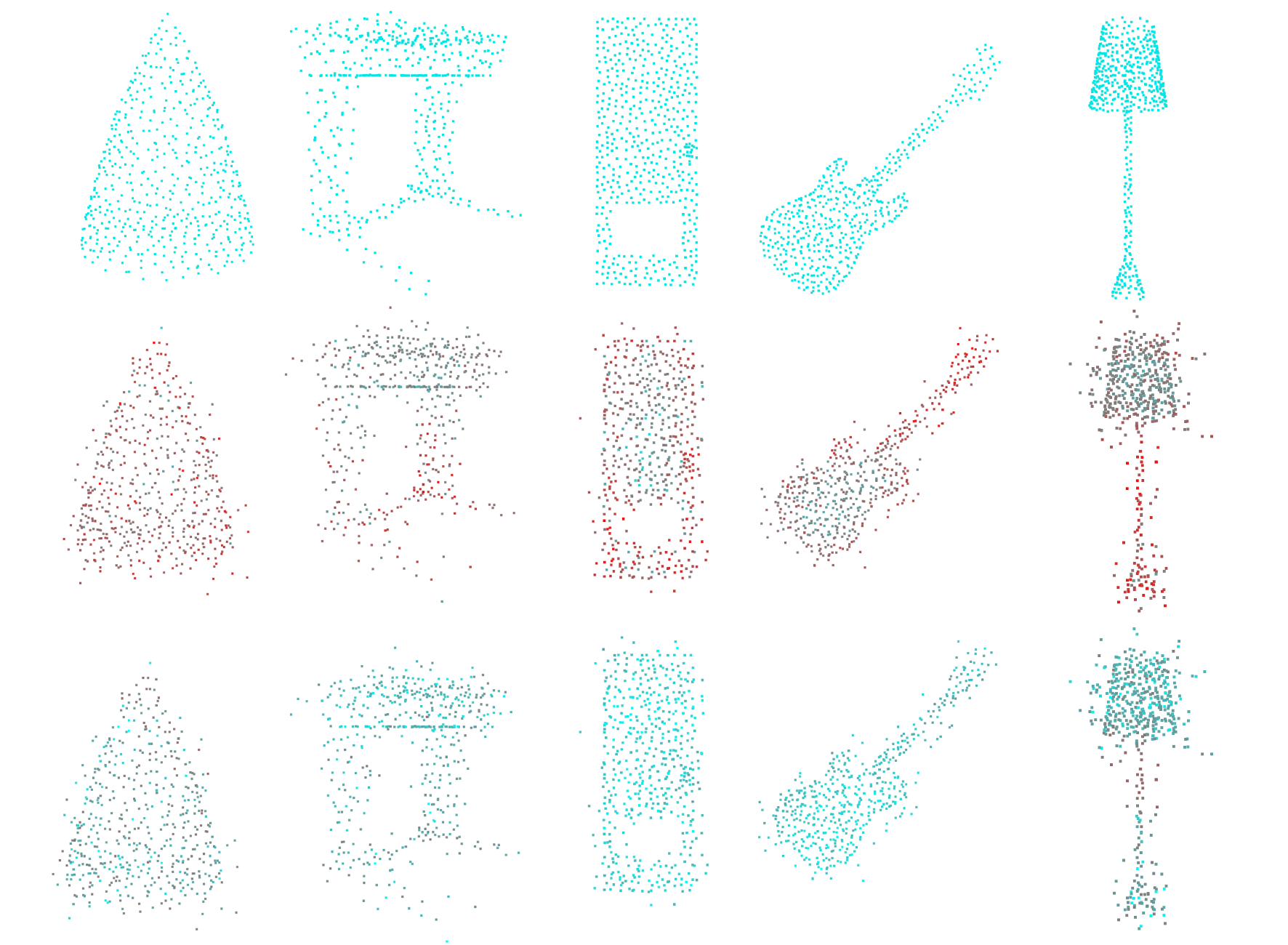}
\end{center}
  \caption{
  Visualization of feature change magnitudes compared to clean input. Blue indicates no change, while red indicates a high degree of change. The graphs in the top row represent the results for a clean sample, the second row displays graphs from PointNet++, and the third row exhibits graphs from Set-Mixer. The noise type is Impulse, and the severity is set to 3.
  }
\label{fig:vis_page3}
\end{figure*}

To demonstrate the impact of noisy points on the extracted features, we employ visualizations to depict the degree of variation in the second layer features of both Set-Mixer and PointNet++.
As illustrated in Fig.~\ref{fig:feature}, when subjected to noise with severity 3, PointNet++ exhibits a significantly greater degree of feature variation.
We further analyze the differences across five noise types on a randomly selected plane sample and a chair sample, as illustrated in Fig.~\ref{fig:vis_page1} and Fig.~\ref{fig:vis_page2}.
Moreover, we visualize samples from more categories, introducing Impulse noise at severity 3, illustrated in Fig.~\ref{fig:vis_page3}.

\end{document}